\documentclass[conference]{IEEEtran}
\IEEEoverridecommandlockouts

\usepackage{cite}
\usepackage{amsmath,amssymb,amsfonts}
\usepackage{algorithmic}
\usepackage{graphicx}
\usepackage{textcomp}
\usepackage{xcolor}
\usepackage{pifont} 
\usepackage{multirow}
\usepackage{colortbl}
\usepackage[numbers]{natbib}         
\usepackage{cite}
\usepackage[colorlinks,
            linkcolor=blue,
            citecolor=blue,
            urlcolor=blue]{hyperref}
\def\BibTeX{{\rm B\kern-.05em{\sc i\kern-.025em b}\kern-.08em
    T\kern-.1667em\lower.7ex\hbox{E}\kern-.125emX}}
\begin{document}

\title{TASAM: Terrain-and-Aware Segment Anything Model for Temporal-Scale Remote Sensing Segmentation\\}
\author{
\IEEEauthorblockN{1\textsuperscript{st} Tianyang Wang}
\IEEEauthorblockA{\textit{University of Alabama at Birmingham}\\
Birmingham, United States\\
toseattle@siu.edu}
\and
\IEEEauthorblockN{2\textsuperscript{nd} Xi Xiao}
\IEEEauthorblockA{\textit{University of Alabama at Birmingham}\\
Birmingham, United States\\
xxiao@uab.edu}
\and
\IEEEauthorblockN{3\textsuperscript{rd} Gaofei Chen}
\IEEEauthorblockA{\textit{University of Alabama at Birmingham}\\
Birmingham, United States\\
gchen2@uab.edu}
\and
\IEEEauthorblockN{4\textsuperscript{th} Hanzhang Chi}
\IEEEauthorblockA{\textit{University of Alabama at Birmingham}\\
Birmingham, United States\\
chihanzhang619@gmail.com}
\and
\IEEEauthorblockN{5\textsuperscript{th} Qi Zhang}
\IEEEauthorblockA{\textit{Wuhan University}\\
Wuhan, China\\
zhangqi\_whursgcm@whu.edu.cn}
\and
\IEEEauthorblockN{6\textsuperscript{th} Guo Cheng}
\IEEEauthorblockA{\textit{Dalian University of Technology}\\
Dalian, China\\
2013012145@dlut.edu.cn}
\and
\IEEEauthorblockN{7\textsuperscript{th} Yingrui Ji\textsuperscript{\dag}}
\IEEEauthorblockA{\textit{University of Chinese Academy of Sciences}\\
Beijing, China\\
jiyingrui1996@gmail.com}
\thanks{\dag\;Corresponding author.}
}


\maketitle

\begin{abstract}
Segment Anything Model (SAM) has demonstrated impressive zero-shot segmentation capabilities across natural image domains, but it struggles to generalize to the unique challenges of remote sensing data, such as complex terrain, multi-scale objects, and temporal dynamics. In this paper, we introduce \textbf{TASAM}, a terrain- and temporally-aware extension of SAM designed specifically for high-resolution remote sensing image segmentation. TASAM integrates three lightweight yet effective modules: a terrain-aware adapter that injects elevation priors, a temporal prompt generator that captures land-cover changes over time, and a multi-scale fusion strategy that enhances fine-grained object delineation. Without retraining the SAM backbone, our approach achieves substantial performance gains across three remote sensing benchmarks—LoveDA, iSAID, and WHU-CD—outperforming both zero-shot SAM and task-specific models with minimal computational overhead. Our results highlight the value of domain-adaptive augmentation for foundation models and offer a scalable path toward more robust geospatial segmentation.
\end{abstract}


\section{Introduction}

Remote sensing image segmentation plays a pivotal role in earth observation, land-use monitoring, disaster assessment, and precision agriculture. With the increasing availability of high-resolution satellite imagery and digital elevation models (DEMs), there is a growing demand for automated, accurate, and generalizable segmentation algorithms that can adapt to complex geographic contexts and temporal variations. Recently, the Segment Anything Model (SAM)~\cite{kirillov2023segment} has emerged as a breakthrough in general-purpose segmentation, offering strong performance across diverse visual domains without task-specific fine-tuning. However, directly applying SAM to remote sensing scenarios reveals a critical performance gap due to three fundamental limitations.

First, remote sensing imagery often contains complex terrain features such as mountainous areas, river valleys, and dense urban clusters that lack prominent RGB contrast but exhibit rich elevation variance. Vanilla SAM, which relies solely on RGB inputs and lacks explicit terrain-awareness, fails to capture these topographical cues, leading to inaccurate boundary detection and under-segmentation in critical regions.

Second, many remote sensing tasks require analyzing temporal dynamics, such as vegetation phenology, urban expansion, and flood progression. However, SAM is inherently static, operating on single-frame inputs without temporal context. This restricts its capacity to leverage the temporal cues essential for detecting gradual or seasonal changes in land cover.

Third, SAM heavily depends on prompt engineering (e.g., box, point, or mask prompts), which is often infeasible or inconsistent in large-scale, unlabeled remote sensing datasets. Moreover, existing prompt formats are not optimized for multi-scale structures or dynamic changes specific to earth observation data.

To address these challenges, we propose \textbf{TASAM} (\textbf{T}errain-\textbf{A}ware \textbf{S}egment \textbf{A}nything \textbf{M}odel), a novel framework that enhances the adaptability of SAM for remote sensing segmentation through multi-modal and multi-temporal conditioning. Our method introduces three core components:
(1)\text{Terrain-Aware Adapter (TA-Adapter):} A lightweight adapter network that injects topographic priors from DEMs into the SAM image encoder, enabling terrain-sensitive segmentation in structurally complex regions. (2)\textbf{Temporal Prompt Generator (TP-Prompt):} A prompt synthesis module that dynamically generates spatial prompts by modeling temporal evolution from multi-date satellite observations, improving change-sensitive segmentation. (3)\textbf{Multi-Scale SAM Fusion (MS-SAM):} A hierarchical token fusion strategy that integrates cross-scale representations within the SAM pipeline, enhancing the segmentation of small objects and fine-grained boundaries.

We conduct comprehensive experiments on three benchmark remote sensing datasets (\textit{LoveDA}, \textit{iSAID}, and \textit{WHU-CD}) and demonstrate that TASAM significantly outperforms both the zero-shot SAM and fine-tuned segmentation baselines, particularly in topographically challenging or temporally dynamic environments.

This paper makes the following contributions:
\begin{itemize}
    \item We identify and formalize three key limitations of applying SAM in remote sensing segmentation: terrain insensitivity, temporal rigidity, and prompt dependence.
    \item We introduce TASAM, a terrain- and temporally-aware extension of SAM, incorporating topographic priors, temporal prompt generation, and multi-scale token fusion.
    \item We empirically validate TASAM on multiple datasets and show substantial improvements over both traditional and recent segmentation baselines under challenging remote sensing scenarios.
\end{itemize}

\section{Related Work}

\subsection{Remote Sensing Image Segmentation}
Remote sensing segmentation underpins land cover classification~\cite{zhu2017deep}, urban mapping~\cite{yuan2018urban}, and disaster monitoring~\cite{chen2020deep}. CNN-based models like U-Net~\cite{ronneberger2015u} and DeepLabV3+~\cite{chen2018encoder} are widely used, while transformers such as Swin-Unet~\cite{cao2021swin} and SegFormer~\cite{xie2021segformer} improve long-range context. Yet these methods require full supervision, remain task-specific, and generalize poorly across domains, often ignoring priors like elevation or temporal change. Our approach instead adapts the general-purpose SAM with terrain- and temporal-aware modules for better domain transfer.

\subsection{Foundation Models and Segment Anything}
Foundation models enable zero/few-shot transfer across vision tasks. CLIP~\cite{radford2021learning, ji2025cibrcrossmodalinformationbottleneck, zhang2025dpcoredynamicpromptcoreset}, DINO~\cite{zhang2022dino}, and SAM~\cite{kirillov2023segment} achieve strong performance in classification, retrieval, and segmentation. SAM employs prompt-driven ViT backbones~\cite{xiao2025roadbenchvisionlanguagefoundationmodel, xiao2025focusfusedobservationchannels, xiao2025visualinstanceawareprompttuning}, but struggles with terrain reasoning, temporal dynamics, and small or low-contrast objects in satellite imagery. While adaptations to medical~\cite{ma2023segment}, agricultural~\cite{sishodia2020app}, and urban~\cite{yu2023urban} domains exist, they remain superficial. We address these bottlenecks via modular extensions injecting multi-source priors.

\subsection{Terrain Modeling and Temporal Change Detection}
Geospatial priors improve segmentation, with DEMs and slope maps aiding landform classification~\cite{li2020deep}, hydrological modeling~\cite{gao2018terrain}, and edge enhancement~\cite{wang2025sopseg}. Temporal fusion is crucial for monitoring land use, vegetation, and infrastructure~\cite{daudt2018urban}. Existing methods often use Siamese~\cite{daudt2018fully} or bi-temporal networks~\cite{ding2022bi}, but these remain incompatible with prompt-driven segmentation. Our framework bridges this gap through terrain- and temporal-aware prompt integration.

\section{Methodology}

\begin{figure}[t]
    \centering
    \includegraphics[width=0.5\textwidth]{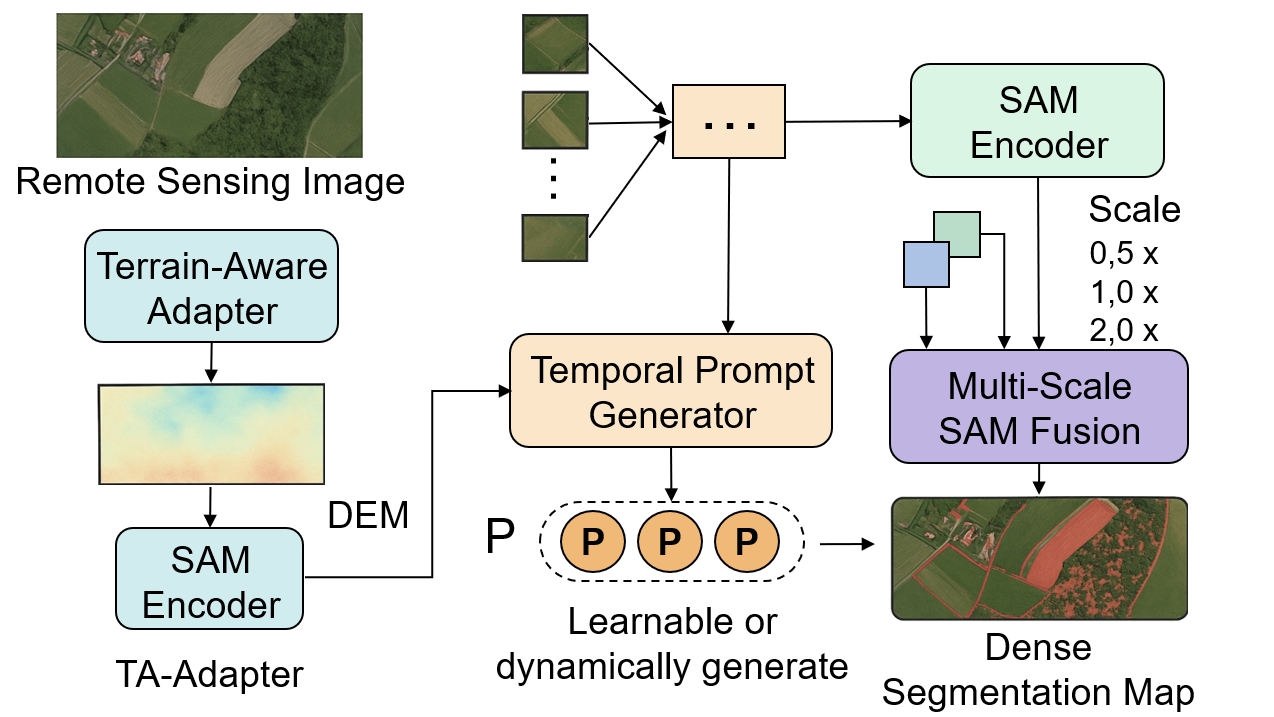}
    \caption{
        Overview of the proposed \textbf{TASAM} framework. Given a remote sensing image, a corresponding elevation map (DEM), and a set of multi-temporal observations, TASAM enhances the Segment Anything Model (SAM) via three key modules: (1) a \textit{Terrain-Aware Adapter} that injects topographic priors into the image encoder, (2) a \textit{Temporal Prompt Generator} that generates spatially adaptive prompts based on temporal dynamics, and (3) a \textit{Multi-Scale SAM Fusion} mechanism that enables robust segmentation across varied object sizes. These extensions allow TASAM to effectively segment complex land-cover scenes without full model fine-tuning.
    }
    \label{fig:architecture}
\end{figure}

We aim to predict a dense semantic map $\hat{Y} \in \mathbb{R}^{H \times W}$ for a satellite image $I \in \mathbb{R}^{H \times W \times 3}$, aided by a DEM $E$ and multi-temporal inputs $\{I_t\}_{t=1}^T$. The task is formulated as learning an augmentation $\mathcal{F}_\theta$ over the frozen SAM encoder $\mathcal{S}$ such that  
\[
\hat{Y} = \mathcal{D}\big(\mathcal{F}_\theta(I,E,\{I_t\}), P\big),
\]
where $\mathcal{D}$ is the SAM mask decoder and $P \in \mathbb{R}^{k \times d}$ denotes learnable or dynamic prompts. As shown in Fig.~\ref{fig:architecture}, our TASAM framework enhances SAM through three lightweight modules. First, the \textit{Terrain-Aware Adapter (TA-Adapter)} injects elevation priors: DEM $E$ is encoded via a CNN into $F_E=\phi_{\mathrm{DEM}}(E)$, then fused with SAM features $F_I=\mathcal{S}_{\mathrm{enc}}(I)$ by gated mixing
\[
F_{\mathrm{TA}} = \gamma \odot F_E + (1-\gamma)\odot F_I,\quad 
\gamma=\sigma(W_\gamma [F_E \| F_I]).
\]
Second, the \textit{Temporal Prompt Generator (TP-Prompt)} encodes historical features $F_t=\mathcal{S}_{\mathrm{enc}}(I_t)$ via self-attention, aggregates them into $F_{\mathrm{temp}}$, and synthesizes $k$ prompts with an MLP:  
\[
P=\mathrm{MLP}(\mathrm{Pool}(F_{\mathrm{temp}})).
\]
Finally, the \textit{Multi-Scale SAM Fusion (MS-SAM)} resizes $I$ into $\{I^{(s)}\}$, extracts features $F^{(s)}=\mathcal{S}_{\mathrm{enc}}(I^{(s)})$, and fuses them by cross-attention:  
\[
F_{\mathrm{MS}}=\mathrm{CrossAttn}(F^{(1)},F^{(2)},F^{(3)}),\quad 
\hat{Y}=\mathcal{D}([F_{\mathrm{MS}} \| P]).
\]
These modules operate plug-and-play, preserve SAM’s frozen backbone, and enable robust segmentation under terrain variation, temporal dynamics, and scale diversity.

\section{Experiments}

\subsection{Experimental Setup}

\paragraph{Datasets} We evaluate TASAM on three remote sensing benchmarks: LoveDA~\cite{wang2021loveda} (8-class land-cover with 598 training and 166 validation images), iSAID~\cite{waqas2019isaid} (655k object instances over 15 categories for dense aerial instance segmentation), and WHU-CD~\cite{ji2018fully} (building change detection for temporal variation analysis).

\paragraph{Evaluation Metrics} Following common practice, we report mean Intersection over Union (mIoU) as the main segmentation metric, along with F1, Precision, and Recall for balance and completeness. Computational efficiency is measured using FLOPs and inference time.

\paragraph{Implementation Details.} All experiments are conducted on an NVIDIA A100 GPU with PyTorch 2.0. We use the official SAM ViT-B model~\cite{kirillov2023segment} as the backbone. During training, we freeze the original SAM encoder and only optimize the introduced modules (adapters, temporal prompt generator, and cross-scale fusion). We use AdamW optimizer with a learning rate of $5 \times 10^{-4}$, batch size of 16, and train for 100 epochs. Multi-scale inputs are generated using image rescaling at factors $\{0.5\times, 1.0\times, 2.0\times\}$, and temporal prompt inputs use $T=3$ frames (e.g., three timestamps for each spatial region). For fair comparison, all baselines are re-trained or evaluated using the same data splits and input resolutions.

\subsection{Comparison with State-of-the-Art Methods}
As shown in Table~\ref{tab:sota_comparison}, TASAM achieves the best mIoU, F1, and recall on LoveDA, iSAID, and WHU-CD, surpassing both general-purpose (SAM) and domain-specific (SegFormer, iFormer) baselines. Fine-tuned SAM narrows the gap but struggles with spatial complexity, whereas TASAM shows clear advantages, especially on WHU-CD (+6.5 mIoU), highlighting the benefit of multi-temporal prompts for change-sensitive segmentation. Table~\ref{tab:subset_analysis} shows TASAM’s robustness in dense urban, rural, and seasonal farmland regions. Terrain- and topography-aware fusion improve boundary delineation, while multi-scale temporal prompting yields over 10 mIoU gains on small-object clusters.

\begin{table}[htbp]
\centering
\caption{\textbf{Comparison with state-of-the-art segmentation methods on three remote sensing benchmarks.} TASAM consistently outperforms both domain-specific baselines across all datasets. Bold numbers indicate the best performance. \textbf{Note:} Prec. presents Precision. SAM (ZS) denotes SAM (Zero-Shot) and SAM (FT) denotes SAM (Fine-Tuned).}
\label{tab:sota_comparison}
\renewcommand{\arraystretch}{1.2}
\setlength{\tabcolsep}{1.5pt}
\begin{tabular}{lcccccc}
\hline
\rowcolor[HTML]{F2F2F2}
\textbf{Dataset} & \textbf{Method} & \textbf{mIoU↑} & \textbf{F1\_Score↑} & \textbf{Prec.↑} & \textbf{Recall↑} & \textbf{Backbone} \\

\hline
\multirow{5}{*}{LoveDA} 
& SAM (ZS)     & 48.3 & 61.0 & 59.2 & 63.1 & ViT-B \\
& SegFormer-B1        & 59.7 & 71.5 & 70.8 & 72.2 & MiT-B1 \\
& iFormer             & 62.0 & 73.1 & 72.4 & 73.8 & iFormer-L \\
& SAM (FT)    & 60.5 & 72.0 & 71.0 & 73.2 & ViT-B \\
& \textbf{TASAM (Ours)} & \textbf{66.3} & \textbf{77.4} & \textbf{76.1} & \textbf{78.8} & ViT-B + Ours \\

\hline
\multirow{5}{*}{iSAID} 
& SAM (ZS)     & 42.8 & 55.6 & 53.7 & 57.5 & ViT-B \\
& SegFormer-B1        & 54.1 & 66.3 & 64.9 & 67.8 & MiT-B1 \\
& iFormer             & 57.9 & 69.2 & 68.0 & 70.3 & iFormer-L \\
& SAM (FT)    & 56.4 & 68.1 & 67.0 & 69.5 & ViT-B \\
& \textbf{TASAM (Ours)} & \textbf{63.7} & \textbf{74.5} & \textbf{73.2} & \textbf{75.9} & ViT-B + Ours \\

\hline
\multirow{5}{*}{WHU-CD} 
& SAM (ZS)     & 51.2 & 63.9 & 62.7 & 65.1 & ViT-B \\
& SegFormer-B1        & 61.5 & 74.3 & 73.1 & 75.6 & MiT-B1 \\
& iFormer             & 64.4 & 76.0 & 75.1 & 77.2 & iFormer-L \\
& SAM (FT)    & 63.3 & 75.1 & 74.0 & 76.4 & ViT-B \\
& \textbf{TASAM (Ours)} & \textbf{69.8} & \textbf{80.5} & \textbf{79.3} & \textbf{81.8} & ViT-B + Ours \\
\hline
\end{tabular}
\end{table}

\begin{figure}[t]
    \centering
    \includegraphics[width=0.48\textwidth]{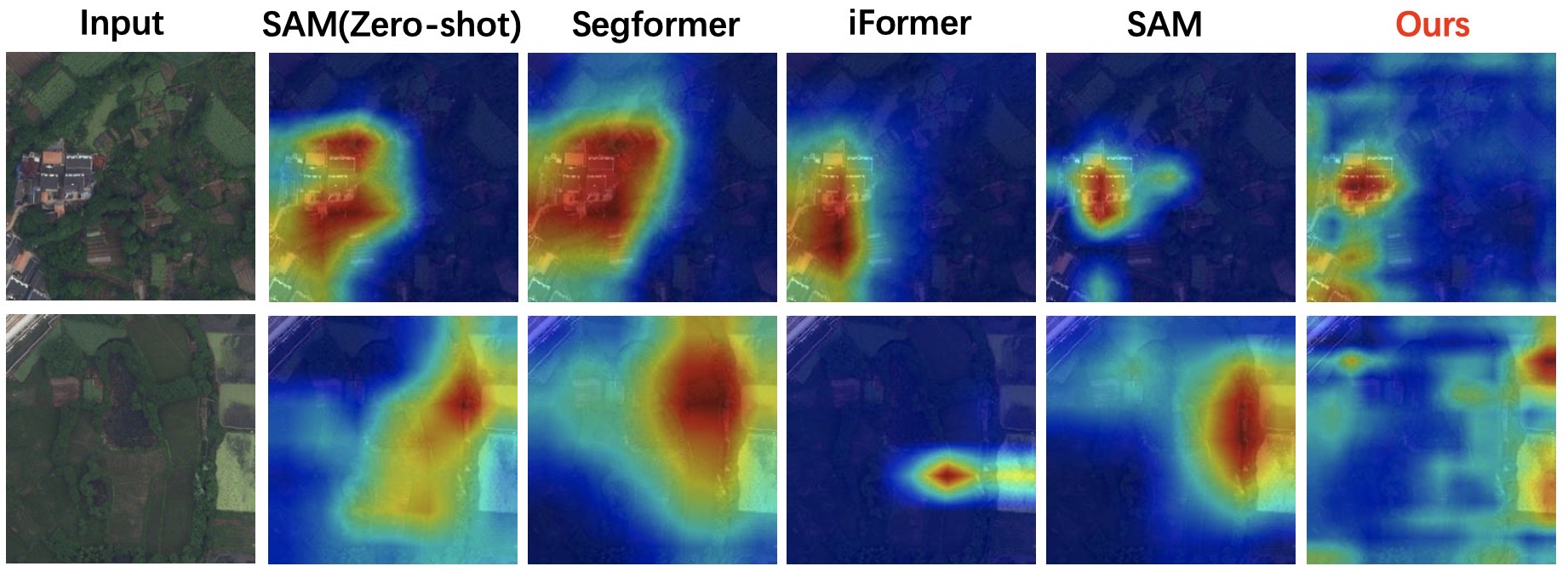}
    \caption{
        \textbf{Visualization of attention heatmaps across methods.} 
        We compare attention maps generated by different segmentation models: SAM (Zero-Shot), Segformer, iFormer, SAM (Fine-Tuned), and our proposed TASAM. 
        Compared to baselines, TASAM exhibits more compact and semantically aligned activations, 
        particularly in rooftop and boundary regions. This suggests improved spatial precision and robustness to clutter.
    }
    
    \label{fig:attention_comparison}
\end{figure}

\begin{table}[t]
\centering
\caption{\textbf{Performance comparison on challenging remote sensing sub-regions.} TASAM demonstrates superior segmentation accuracy on small objects, complex terrain, and temporally dynamic regions. Bold values indicate the best results.}
\label{tab:subset_analysis}
\renewcommand{\arraystretch}{1.2}
\setlength{\tabcolsep}{2.5pt}
\begin{tabular}{lcccc}
\hline
\rowcolor[HTML]{F2F2F2}
\textbf{Scene Type} & \textbf{Method} & \textbf{mIoU ↑} & \textbf{Precision ↑} & \textbf{Recall ↑} \\
\hline

\multirow{4}{*}{\begin{tabular}{c}Urban High-Rise \\ (Dense Buildings)\end{tabular}}
& SAM (Zero-Shot)     & 45.2 & 51.0 & 49.5 \\
& SAM (Fine-Tuned)    & 53.9 & 62.1 & 58.2 \\
& iFormer             & 58.3 & 66.4 & 62.0 \\
& \textbf{TASAM (Ours)} & \textbf{63.7} & \textbf{72.8} & \textbf{67.4} \\

\hline

\multirow{4}{*}{\begin{tabular}{c}Rural Terrain \\ (Hill/Valley)\end{tabular}}
& SAM (Zero-Shot)     & 44.1 & 50.2 & 48.0 \\
& SAM (Fine-Tuned)    & 52.5 & 59.3 & 56.1 \\
& iFormer             & 55.7 & 63.8 & 60.4 \\
& \textbf{TASAM (Ours)} & \textbf{61.0} & \textbf{69.1} & \textbf{65.7} \\

\hline

\multirow{4}{*}{\begin{tabular}{c}Seasonal Farmland \\ (Time-Variant)\end{tabular}}
& SAM (Zero-Shot)     & 46.6 & 54.5 & 51.3 \\
& SAM (Fine-Tuned)    & 55.4 & 63.1 & 61.2 \\
& iFormer             & 57.9 & 66.2 & 64.3 \\
& \textbf{TASAM (Ours)} & \textbf{65.6} & \textbf{73.5} & \textbf{71.4} \\

\hline

\multirow{4}{*}{\begin{tabular}{c}Small Object Cluster \\ (Vehicles, Rooftops)\end{tabular}}
& SAM (Zero-Shot)     & 39.3 & 44.8 & 42.0 \\
& SAM (Fine-Tuned)    & 48.7 & 54.2 & 51.5 \\
& iFormer             & 52.0 & 58.6 & 56.0 \\
& \textbf{TASAM (Ours)} & \textbf{58.9} & \textbf{65.7} & \textbf{62.3} \\
\hline
\end{tabular}
\end{table}

\subsection{Ablation Study}

We conduct a comprehensive ablation study to validate the contribution of each core component in TASAM, including the Terrain-Aware Adapter, Temporal Prompt Generator, and Multi-Scale Fusion module. Table~\ref{tab:ablation_loveda} summarizes the mIoU results on the LoveDA dataset for various model variants. Starting from the fine-tuned SAM baseline (60.5 mIoU), each proposed module brings a consistent improvement. Removing the Terrain Adapter drops performance to 62.8, indicating that incorporating elevation priors significantly enhances segmentation in topographically complex regions. Removing the Temporal Prompt module results in a larger drop (63.1 mIoU), suggesting that capturing temporal changes is essential for dynamic or seasonal landscapes. The Multi-Scale Fusion module also proves important, as omitting it leads to a 2.1 point drop. The full TASAM model achieves 66.3 mIoU.

\begin{figure}[htbp]
    \centering
    \includegraphics[width=0.8\linewidth]{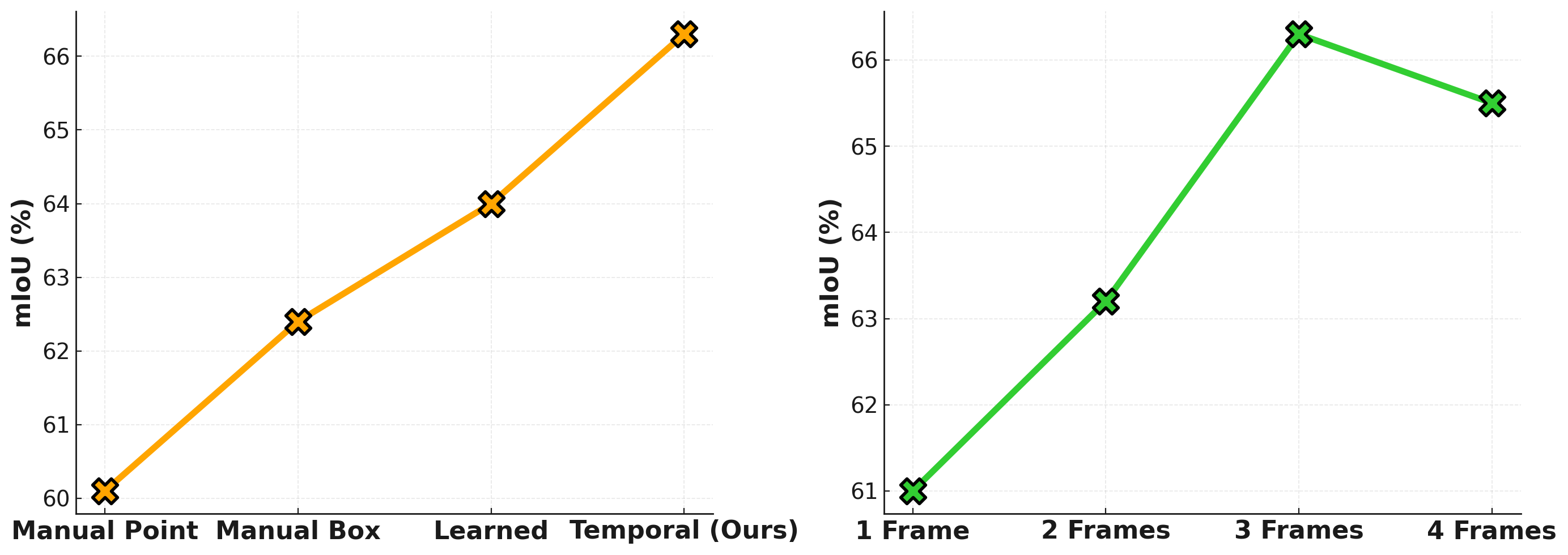}
    \caption{\textbf{Prompt strategy and temporal window ablation.} 
    Left: Our temporal prompt mechanism outperforms manual point/box and learned prompts.
    Right: Increasing the temporal window size improves performance up to 3 frames, after which performance plateaus or drops.}
    \label{fig:ablation_prompt_temporal}
\end{figure}

Figure~\ref{fig:ablation_prompt_temporal} provides additional insights. The left plot compares different prompt strategies: manually annotated points and boxes yield lower accuracy due to sparse spatial coverage, while learned prompts offer modest improvement. The right plot shows how performance varies with the temporal window size: while using more frames improves accuracy up to 3 frames, adding more leads to marginal degradation, likely due to increased noise and redundancy.

\subsection{Qualitative Results and Visualization}

To further investigate the spatial reasoning behavior of different models, we visualize attention heatmaps in Figure~\ref{fig:attention_comparison}. Compared to SAM (Zero-Shot), Segformer, iFormer, and even fine-tuned SAM, our TASAM produces attention maps that are both more compact and more semantically aligned with actual object boundaries. In complex rural and urban scenes, TASAM effectively suppresses background noise and focuses on topographically relevant regions—e.g., rooftops and roads—while avoiding diffuse or fragmented attention. The sharper focus and reduced false activations demonstrate the benefit of integrating terrain priors and temporal context into prompt-aware segmentation. These visualizations not only confirm the quantitative improvements but also highlight the interpretability and robustness of TASAM in practical scenarios.

\begin{table}[htbp]
\centering
\caption{\textbf{Ablation study on LoveDA.} Each module contributes to mIoU improvement.}
\label{tab:ablation_loveda}
\setlength{\tabcolsep}{5.5pt}
\begin{tabular}{lccc|c}
\hline
\rowcolor[HTML]{F2F2F2}
\textbf{Variant} & \textbf{Terrain} & \textbf{Temporal} & \textbf{Multi-Scale} & \textbf{mIoU ↑} \\
\hline
SAM (Fine-Tuned)              & \ding{55} & \ding{55} & \ding{55} & 60.5 \\
w/o Terrain Adapter           & \ding{55} & \ding{51} & \ding{51} & 62.8 \\
w/o Temporal Prompt           & \ding{51} & \ding{55} & \ding{51} & 63.1 \\
w/o Multi-Scale Fusion        & \ding{51} & \ding{51} & \ding{55} & 64.2 \\
\textbf{TASAM (Full)}         & \ding{51} & \ding{51} & \ding{51} & \textbf{66.3} \\
\hline
\end{tabular}
\end{table}


\subsection{Prompt Sensitivity and Generalization}

We first examine the impact of prompt quantity on segmentation performance (Table~\ref{tab:prompt_quantity}). TASAM achieves optimal results using four prompts, while further increasing the number leads to saturation or slight decline. This indicates that overly dense prompting introduces redundancy without contributing meaningful guidance. 

\begin{table}[t]
\centering
\caption{\textbf{Effect of prompt quantity on segmentation performance.} 
TASAM achieves the best results with 4 prompts. Performance saturates or slightly drops beyond this point due to potential redundancy.}
\label{tab:prompt_quantity}
\setlength{\tabcolsep}{8pt}
\renewcommand{\arraystretch}{1.15}
\begin{tabular}{c|ccccc}
\hline
\rowcolor[HTML]{F2F2F2}
\textbf{\#Prompts} & \textbf{1} & \textbf{2} & \textbf{4} & \textbf{6} & \textbf{8} \\
\hline
mIoU (\%)     & 63.2 & 64.4 & \textbf{66.3} & 66.2 & 65.9 \\
F1 Score (\%) & 74.6 & 75.9 & \textbf{77.4} & 77.2 & 76.8 \\
Recall (\%)   & 75.2 & 76.5 & \textbf{78.8} & 78.4 & 78.0 \\
\hline
\end{tabular}
\end{table}

\subsection{Efficiency and Scalability}

We report the model complexity and inference efficiency of TASAM in Table~\ref{tab:efficiency_comparison}. While TASAM introduces only a modest increase in parameters (+1.6M) and FLOPs (+2.4G) over SAM (ViT-B), it achieves a substantial improvement of +5.8\% mIoU. 
Moreover, TASAM maintains real-time inference capability, with only a 4.4ms increase in per-image runtime compared to the SAM baseline. When compared to iFormer, which has fewer parameters but lower accuracy, TASAM offers a more balanced trade-off between accuracy and computational cost. 

\begin{table}[htbp]
\centering
\caption{\textbf{Efficiency and scalability comparison.} 
TASAM adds minimal overhead over SAM while improving performance.}
\label{tab:efficiency_comparison}
\renewcommand{\arraystretch}{1.15}
\setlength{\tabcolsep}{4.5pt}
\small
\begin{tabular}{l|cccc}
\rowcolor[HTML]{F2F2F2}
\hline
\textbf{Method} & \textbf{mIoU (\%)} &
\textbf{Params} & \textbf{FLOPs} & \textbf{Time} \\
               &                    & (M)             & (G)            & (ms)          \\
\hline
SAM (ViT-B)         & 60.5 & 86.0  & 89.7 & 132.4 \\
iFormer             & 62.0 & 72.8  & 65.3 & 101.6 \\
\textbf{TASAM (Ours)} & \textbf{66.3} & 87.6  & 92.1 & 136.8 \\
\hline
\end{tabular}
\end{table}




\section{Conclusion}

In this paper, we presented \textbf{TASAM}, a terrain- and temporally-aware extension of the Segment Anything Model tailored for remote sensing image segmentation. By incorporating digital elevation priors, temporal prompt generation, and multi-scale fusion into the SAM architecture, TASAM addresses key challenges in geographic complexity, temporal dynamics, and scale variability. Extensive experiments on multiple remote sensing benchmarks demonstrate that TASAM significantly outperforms both foundation models and domain-specific baselines, while introducing only minimal computational overhead. Our findings suggest that domain-adaptive augmentation of foundation models is a promising direction for geospatial understanding. In future work, we plan to explore incorporating semantic priors from multispectral or hyperspectral modalities and extending TASAM to instance-level and 3D segmentation tasks.





\bibliographystyle{IEEEtran}
\bibliography{IEEE-conference-template-062824}

\begin{thebibliography}{10}
\providecommand{\url}[1]{#1}
\csname url@samestyle\endcsname
\providecommand{\newblock}{\relax}
\providecommand{\bibinfo}[2]{#2}
\providecommand{\BIBentrySTDinterwordspacing}{\spaceskip=0pt\relax}
\providecommand{\BIBentryALTinterwordstretchfactor}{4}
\providecommand{\BIBentryALTinterwordspacing}{\spaceskip=\fontdimen2\font plus
\BIBentryALTinterwordstretchfactor\fontdimen3\font minus \fontdimen4\font\relax}
\providecommand{\BIBforeignlanguage}[2]{{%
\expandafter\ifx\csname l@#1\endcsname\relax
\typeout{** WARNING: IEEEtran.bst: No hyphenation pattern has been}%
\typeout{** loaded for the language `#1'. Using the pattern for}%
\typeout{** the default language instead.}%
\else
\language=\csname l@#1\endcsname
\fi
#2}}
\providecommand{\BIBdecl}{\relax}
\BIBdecl

\bibitem{kirillov2023segment}
A.~Kirillov, E.~Mintun, N.~Ravi, H.~Mao, C.~Rolland, L.~Gustafson, T.~Xiao, S.~Whitehead, A.~C. Berg, W.-Y. Lo \emph{et~al.}, ``Segment anything,'' in \emph{ICCV}, 2023, pp. 4015--4026.

\bibitem{zhu2017deep}
X.~X. Zhu, D.~Tuia, L.~Mou, G.-S. Xia, L.~Zhang, F.~Xu, and F.~Fraundorfer, ``Deep learning in remote sensing: A comprehensive review and list of resources,'' \emph{IEEE GRS}, vol.~5, no.~4, pp. 8--36, 2017.

\bibitem{yuan2018urban}
C.~Yuan, R.~Shan, Y.~Zhang, X.-X. Li, T.~Yin, J.~Hang, and L.~Norford, ``Multilayer urban canopy modelling and mapping for traffic pollutant dispersion at high density urban areas,'' \emph{Science of the total environment}, vol. 647, pp. 255--267, 2019.

\bibitem{chen2020deep}
D.~Chen, Z.~Liu, L.~Wang, M.~Dou, J.~Chen, and H.~Li, ``Natural disaster monitoring with wireless sensor networks: A case study of data-intensive applications upon low-cost scalable systems,'' \emph{Mobile Networks and Applications}, vol.~18, no.~5, pp. 651--663, 2013.

\bibitem{ronneberger2015u}
O.~Ronneberger, P.~Fischer, and T.~Brox, ``U-net: Convolutional networks for biomedical image segmentation,'' in \emph{MICCAI}.\hskip 1em plus 0.5em minus 0.4em\relax Springer, 2015, pp. 234--241.

\bibitem{chen2018encoder}
L.-C. Chen, Y.~Zhu, G.~Papandreou, F.~Schroff, and H.~Adam, ``Encoder-decoder with atrous separable convolution for semantic image segmentation,'' in \emph{ECCV}, 2018, pp. 801--818.

\bibitem{cao2021swin}
H.~Cao, Y.~Wang, J.~Chen, D.~Jiang, X.~Zhang, Q.~Tian, and M.~Wang, ``Swin-unet: Unet-like pure transformer for medical image segmentation,'' in \emph{ECCV}.\hskip 1em plus 0.5em minus 0.4em\relax Springer, 2022, pp. 205--218.

\bibitem{xie2021segformer}
E.~Xie, W.~Wang, Z.~Yu, A.~Anandkumar, J.~M. Alvarez, and P.~Luo, ``Segformer: Simple and efficient design for semantic segmentation with transformers,'' \emph{NeurIPS}, vol.~34, pp. 12\,077--12\,090, 2021.

\bibitem{radford2021learning}
A.~Radford, J.~W. Kim, C.~Hallacy, A.~Ramesh, G.~Goh, S.~Agarwal, G.~Sastry, A.~Askell, P.~Mishkin, J.~Clark \emph{et~al.}, ``Learning transferable visual models from natural language supervision,'' in \emph{ICML}, 2021, pp. 8748--8763.

\bibitem{ji2025cibrcrossmodalinformationbottleneck}
Y.~Ji, X.~Xiao, G.~Chen, H.~Xu, C.~Ma, L.~Zhu, A.~Liang, and J.~Chen, ``Cibr: Cross-modal information bottleneck regularization for robust clip generalization,'' 2025, arXiv:2503.24182.

\bibitem{zhang2025dpcoredynamicpromptcoreset}
Y.~Zhang, A.~Mehra, S.~Niu, and J.~Hamm, ``Dpcore: Dynamic prompt coreset for continual test-time adaptation,'' 2025.

\bibitem{zhang2022dino}
H.~Zhang, F.~Li, S.~Liu, L.~Zhang, H.~Su, J.~Zhu, L.~M. Ni, and H.-Y. Shum, ``Dino: Detr with improved denoising anchor boxes for end-to-end object detection,'' \emph{arXiv:2203.03605}, 2022.

\bibitem{xiao2025roadbenchvisionlanguagefoundationmodel}
X.~Xiao, Y.~Zhang, J.~Wang, L.~Zhao, Y.~Wei, H.~Li, Y.~Li, X.~Wang, S.~K. Roy, H.~Xu, and T.~Wang, ``Roadbench: A vision-language foundation model and benchmark for road damage understanding,'' 2025, arXiv:2507.17353.

\bibitem{xiao2025focusfusedobservationchannels}
X.~Xiao, A.~Tsaris, A.~Tabassum, J.~Lagergren, L.~M. York, T.~Wang, and X.~Wang, ``Focus: Fused observation of channels for unveiling spectra,'' 2025, arXiv:2507.14787.

\bibitem{xiao2025visualinstanceawareprompttuning}
X.~Xiao, Y.~Zhang, X.~Li, T.~Wang, X.~Wang, Y.~Wei, J.~Hamm, and M.~Xu, ``Visual instance-aware prompt tuning,'' 2025, arXiv:2507.07796.

\bibitem{ma2023segment}
Z.~Ma, X.~He, S.~Sun, B.~Yan, H.~Kwak, and J.~Gao, ``Zero-shot digital rock image segmentation with a fine-tuned segment anything model,'' 2023.

\bibitem{sishodia2020app}
R.~P. Sishodia, R.~L. Ray, and S.~K. Singh, ``Applications of remote sensing in precision agriculture: A review,'' \emph{Remote sensing}, vol.~12, no.~19, p. 3136, 2020.

\bibitem{yu2023urban}
D.~Yu and C.~Fang, ``Urban remote sensing with spatial big data: A review and renewed perspective of urban studies in recent decades,'' \emph{Remote Sensing}, vol.~15, no.~5, p. 1307, 2023.

\bibitem{li2020deep}
J.~Li, Y.~Zhao, P.~Bates, J.~Neal, S.~Tooth, L.~Hawker, and C.~Maffei, ``Digital elevation models for topographic characterisation and flood flow modelling along low-gradient, terminal dryland rivers: A comparison of spaceborne datasets for the r{\'\i}o colorado, bolivia,'' \emph{Journal of Hydrology}, vol. 591, p. 125617, 2020.

\bibitem{gao2018terrain}
J.~Gao, D.~Peng, T.~Zhou, T.~Wang, and C.~Xu, ``Terrain matching localization for underwater vehicle based on gradient fitting,'' \emph{Journal of Sensors}, vol. 2018, no.~1, p. 3717430, 2018.

\bibitem{wang2025sopseg}
C.~Wang, Y.~Ji, Y.~Meng, Y.~Zhang, and Y.~Zhu, ``Sopseg: Prompt-based small object instance segmentation in remote sensing imagery,'' 2025.

\bibitem{daudt2018urban}
R.~C. Daudt, B.~Le~Saux, A.~Boulch, and Y.~Gousseau, ``Urban change detection for multispectral earth observation using convolutional neural networks,'' in \emph{IGARSS 2018}.\hskip 1em plus 0.5em minus 0.4em\relax Ieee, 2018, pp. 2115--2118.

\bibitem{daudt2018fully}
R.~C. Daudt, B.~Le~Saux, and A.~Boulch, ``Fully convolutional siamese networks for change detection,'' in \emph{2018 25th IEEE ICIP}.\hskip 1em plus 0.5em minus 0.4em\relax IEEE, 2018, pp. 4063--4067.

\bibitem{ding2022bi}
L.~Ding, H.~Guo, S.~Liu, L.~Mou, J.~Zhang, and L.~Bruzzone, ``Bi-temporal semantic reasoning for the semantic change detection in hr remote sensing images,'' \emph{IEEE TGRS}, vol.~60, pp. 1--14, 2022.

\bibitem{wang2021loveda}
J.~Wang, Z.~Zheng, A.~Ma, X.~Lu, and Y.~Zhong, ``Loveda: A remote sensing land-cover dataset for domain adaptive semantic segmentation,'' 2021.

\bibitem{waqas2019isaid}
S.~Waqas~Zamir, A.~Arora, A.~Gupta, S.~Khan, G.~Sun, F.~Shahbaz~Khan, F.~Zhu, L.~Shao, G.-S. Xia, and X.~Bai, ``isaid: A large-scale dataset for instance segmentation in aerial images,'' in \emph{CVPR workshops}, 2019, pp. 28--37.

\bibitem{ji2018fully}
S.~Ji, S.~Wei, and M.~Lu, ``Fully convolutional networks for multisource building extraction from an open aerial and satellite imagery data set,'' \emph{IEEE TGRS}, vol.~57, no.~1, pp. 574--586, 2018.

\end{thebibliography}


\end{document}